\documentclass[letterpaper]{article}
\usepackage[preprint]{aaai2027}
\usepackage[hyphens]{url}
\usepackage{graphicx}
\usepackage{natbib}
\usepackage{caption}
\usepackage{amsmath}
\usepackage{amssymb}
\usepackage{amsfonts}
\usepackage{textcomp}
\usepackage{xcolor}
\usepackage{booktabs}
\usepackage{array}

\newcommand{\method}{\textsc{EndoUT}}
\newcommand{\dutility}{\Delta\mathrm{Utility}}

\title{Endogenous Preferences as Vulnerability Signals: Discovering Universal Adversarial Triggers for LLM Utility Degradation}
\author{
Qingyu Wu\textsuperscript{\rm 1}\equalcontrib,
Zeyu Feng\textsuperscript{\rm 1}\equalcontrib,
Yuzhe Luo\textsuperscript{\rm 1},
Hua Cheng\textsuperscript{\rm 1}\corresponding
}
\affiliations{
\textsuperscript{\rm 1}The Defense Innovation Institute, Academy of Military Sciences, Beijing, China
}

\begin{document}

\maketitle

\begin{abstract}
Deployed large language models must remain useful when untrusted text enters their prompt context. Adversarial prompting and red-teaming methods typically define the failures they seek through task labels, harmful targets, judge feedback, or downstream utility scores. We ask whether a frozen model's clean behavior can reveal a reusable context vulnerability when its task-level form is unknown, and downstream task feedback is withheld. To study this question, EndoUT uses a Siamese-style design that compares the same frozen victim on clean and span-modified versions of each unlabeled instruction. Without a task-level target, the clean pass supplies the victim's preferred response-onset token as an instruction-conditioned reference; Endogenous Preference Disruption (EPD) measures how insertion weakens that preference. An auxiliary generator supplies bounded-span proposals; EPD guides its update and ranks the pool without access to downstream information. The highest-scoring victim-specific span is fixed before evaluation and reused unchanged across all held-out inputs. Across four instruction-tuned models and nine benchmarks, fixed triggers reduce benign-task utility by 40.2 percentage points on average, with negative mean change on every benchmark and greater average degradation than every evaluated baseline under the reported protocols. Matched protocol ablations show stronger transfer for source-side selection followed by fixed reuse than for null-context fixed reuse or per-input generation. Transfer recurs across proposers with proposal-dependent strength; output-level analyses identify victim-specific associations between trigger families and downstream failure patterns. Thus, clean behavior can expose context vulnerabilities that transfer across inputs yet manifest through structured, victim-specific patterns of utility degradation. The code and data will be released upon acceptance.
\end{abstract}

\section{Introduction}
Prompt context becomes an attack surface whenever untrusted text is processed alongside a benign request.
The resulting failure need not involve prohibited content: an inserted span may instead disrupt reasoning, answer format, or instruction following and reduce benign utility.
If the same span produces such losses across unrelated requests, it exposes a reusable context vulnerability.
Can the clean behavior of a frozen victim model reveal such a span when discovery neither specifies the downstream form of failure nor receives feedback from downstream tasks?

Adversarial search and robustness evaluation typically begin by specifying either the behavior to induce or the task-level criterion to degrade.
Universal-trigger methods use objectives defined by labels, target classes or tokens, or task losses \citep{wallace2019universal}, while automated jailbreaks reward harmful compliance, refusal bypass, target matching, or judge success \citep{zou2023universal,liu2023autodan,chao2023pair,mehrotra2023tap,yu2023gptfuzzer,paulus2024advprompter}.
Prompt-robustness work examines predefined changes in wording, spelling, punctuation, formatting, or semantics \citep{zhu2023promptbench,cao2024worstprompt,seleznyov2025punctuation,zheng2025promptanatomy}.
Availability-oriented attacks target instruction redirection, guardrail activation, resource exhaustion, or system-channel poisoning \citep{greshake2023indirect,zhang2024safeguarddos,gao2024pdos,li2025systempromptpoisoning}.
These methods make specified failures tractable, but leave open whether the model's clean behavior can guide discovery when no task-level manifestation is specified during search.
Withholding downstream feedback prevents the search objective from defining in advance what counts as task-level failure; utility change on held-out tasks therefore measures transfer rather than direct optimization.

Discovery under this separation still requires a source-side basis for comparing candidate spans.
That basis must be available for every unlabeled instruction, remain conditioned on that instruction, and prescribe no downstream error.
The victim model's preferred token at response onset satisfies these requirements: it is instruction-conditioned yet requires no label, answer key, or task scorer.
Candidate spans can therefore be compared on the same instruction by how strongly each suppresses the probability of that token.
Held-out evaluation then tests whether a span selected by this source-local comparison exposes a reusable vulnerability when fixed and applied across new inputs and tasks.

\begin{figure*}[t]
\centering
\includegraphics[width=0.98\textwidth]{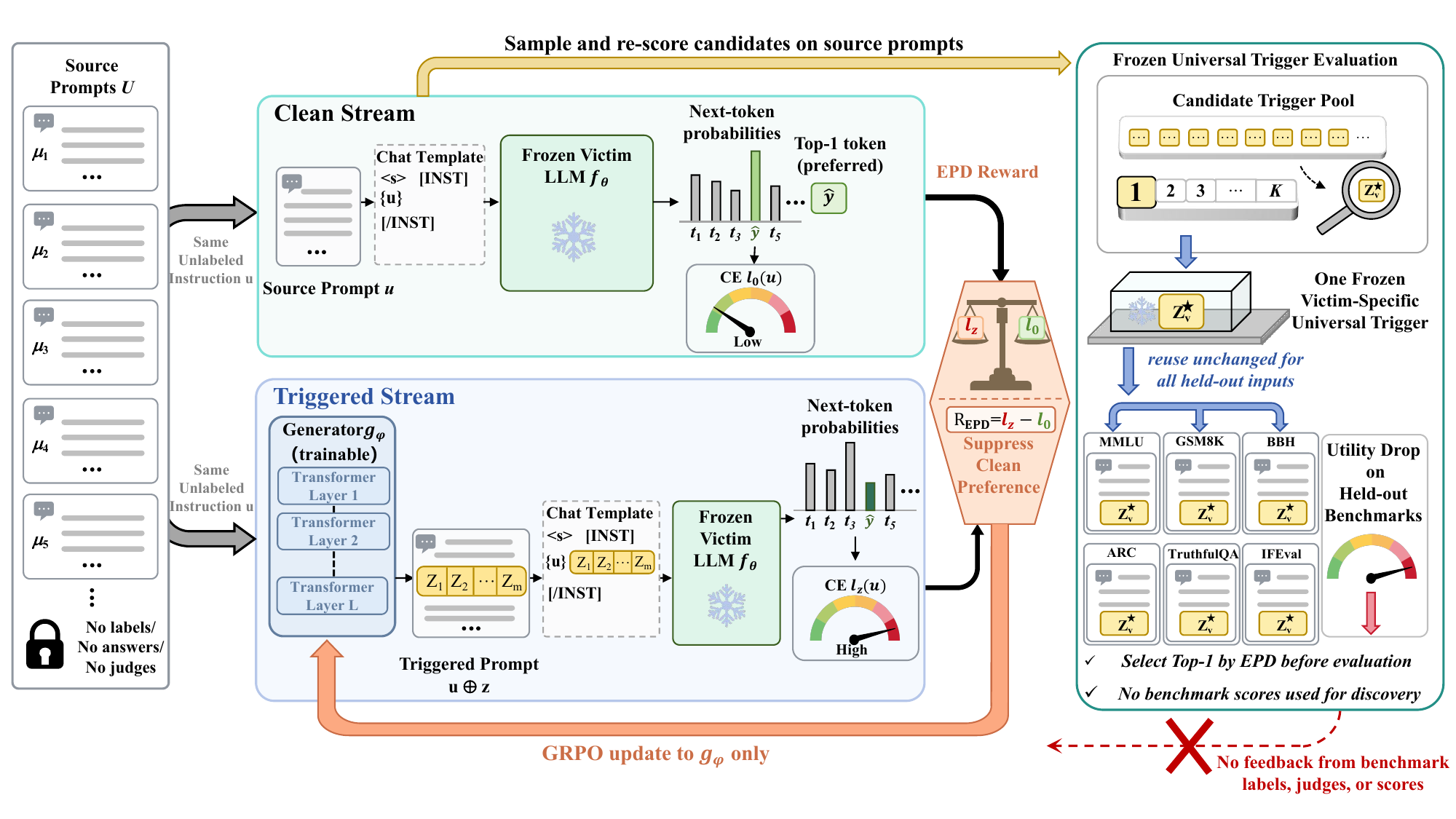}
\vspace{-0.4cm}
\caption{Source-side discovery and held-out transfer in \method{}. A frozen victim's clean response-onset preference provides the reference for scoring bounded spans proposed by an auxiliary generator from unlabeled instructions. One victim-specific span is selected without benchmark feedback, then fixed and reused across held-out inputs and tasks.}
\label{fig:endout_framework}
\end{figure*}

\method{} implements this test by separating source-side discovery from held-out utility measurement (Figure~\ref{fig:endout_framework}).
On each source instruction, Endogenous Preference Disruption (EPD) records the increase in the clean preferred token's negative log-probability after candidate insertion.
To explore the discrete space of bounded spans, an auxiliary generator serves as the proposal mechanism; EPD guides this exploration and ranks the resulting source-side pool, while the victim model remains frozen.
The highest-ranked victim-specific span becomes the fixed trigger before any benchmark information is used and is then reused unchanged across held-out inputs.

Across four instruction-tuned victim models and nine public benchmarks, the fixed triggers reduce utility by 40.2 percentage points on average.
The four-victim mean is negative on every benchmark, and the average reduction exceeds that of every evaluated baseline under the reported protocols.
Transfer also extends beyond a single selected trigger: all 20 source-ranked candidates for each victim have negative mean utility change, as do the fixed triggers from all 24 victim--generator pairs.
Degradation magnitude nevertheless varies across realized proposal pools.
Downstream regressions are structured rather than diffuse: in 19 of the 24 victim--trigger pairs, at least 70\% occupy one victim-local output cluster, while the dominant semantic category differs across the four reviewed main-protocol pairs.
The combined evidence shows that aggregate degradation alone does not characterize the discovered vulnerability: finite proposal search conditions what becomes visible, while downstream expression is structured at the victim--trigger level.

The main contributions are:
\begin{itemize}
\item We formulate endogenous discovery of reusable context vulnerabilities in frozen LLMs: selecting a victim-specific fixed span from unlabeled instructions without specifying downstream failure or consulting downstream task feedback.
\item \method{} operationalizes this formulation through paired clean and span-modified passes of the same frozen victim. Across four models and nine benchmarks, its fixed triggers average utility reduction exceeding every evaluated baseline under the reported protocols and establish the severity of the exposed vulnerability.
\item Empirically, downstream-agnostic discovery uncovers different, recurring failure modes across unrelated tasks, while remaining bounded by the generator's finite search support. The victim specificity of both recovered triggers and their induced failure modes exposes a systematic, model-dependent vulnerability that warrants closer scrutiny.
\end{itemize}

\section{Related Work}

\subsection{Textual Adversarial Attacks}
Textual adversarial attacks optimize discrete perturbations under task-specified objectives, ranging from character- and word-level edits against supervised models to input-agnostic token sequences \citep{jin2019textfooler,li2020bertattack,boucher2021badcharacters,wallace2019universal}.
Subsequent work reduces reliance on labeled examples, studies universal and natural triggers in prompt-based models, and searches hard prompts using target tokens, answer spans, labels, or task losses \citep{parekh2021minimal,xu2022universalvulnerability,xu2024linkprompt,meade2024notuniversal,shin2020autoprompt,wen2023hardprompts}.
Untargeted objectives need not specify a particular wrong output; \method{} further withholds downstream task feedback and measures utility only after a victim-specific span is fixed.

For aligned LLMs, automated attacks search adversarial suffixes, semantic prompts, iterative procedures, fuzzing policies, or learned generators for harmful compliance or refusal bypass \citep{zou2023universal,liu2023autodan,chao2023pair,mehrotra2023tap,yu2023gptfuzzer,liao2024amplegcg,paulus2024advprompter,sadasivan2024beast}.
JailbreakBench, HarmBench, and StrongREJECT standardize these evaluations with harmful-behavior datasets and judges \citep{mazeika2024harmbench}.

\subsection{LLM Utility and Availability Failures}
LLM security also concerns failures that reduce usefulness or responsiveness without eliciting harmful content.
Prompt injection can produce such losses by exploiting instruction hierarchy, untrusted context, or system-prompt channels to redirect model behavior \citep{greshake2023indirect,li2025systempromptpoisoning}.
Other availability-oriented work studies guardrail false positives and poisoning-based denial of service \citep{zhang2024safeguarddos,gao2024pdos}, over-generation and pathological reasoning costs \citep{manu2025overgeneration,liu2026reasoningbomb}, and soft failures in retrieval-augmented systems \citep{zhang2026softfailure}.
Benchmarks for safety--utility tradeoffs further show that protective mechanisms can themselves reduce task performance \citep{mai2025usebench}.

\subsection{Prompt Robustness}
Prompt robustness evaluates whether LLMs preserve task performance under changes to the prompt itself.
PromptBench studies predefined character-, word-, sentence-, and semantic-level prompt perturbations across benchmark tasks \citep{zhu2023promptbench}.
Work on worst-case prompts, punctuation, formatting, and prompt components further documents heterogeneous sensitivity across task formulations \citep{cao2024worstprompt,seleznyov2025punctuation,zheng2025promptanatomy}.

\section{Vulnerability Discovery}
\subsection{Discovery Setting and Threat Model}
We study whether a victim model's clean behavior can reveal a fixed span whose effects transfer across held-out tasks, even when discovery neither specifies the downstream failure form nor receives task-level feedback.
The adversary may place one bounded span in the user-visible prompt context; the victim model remains frozen, and the benign request and downstream task protocol are otherwise unchanged.
One victim-specific span is selected before held-out evaluation; this fixed trigger is then reused without adaptation across new inputs and tasks, and its effect is quantified only afterward as aggregate utility change.
The main protocol appends the trigger as a user-content suffix; prefix and random-position variants test sensitivity to insertion position.

Let $f_\theta$ denote this victim model.
For a benign instruction $x$ and a candidate span $z$ subject to token budget $L$, the main insertion operator appends $z$ to the user content, $T_z(x)=x\oplus z$.
The adversary may query next-token probabilities from $f_\theta$ on unlabeled source instructions.
No externally specified target token or response, held-out benchmark input or label, task scorer, judge feedback, or utility measurement is available during discovery; task-level information enters only after $z$ is fixed.

Following prior LLM prompt-robustness and safety--utility evaluations \citep{zhu2023promptbench,cao2024worstprompt,mai2025usebench}, we treat each benchmark's official normalized score as task utility.
We report utility change as attacked minus clean utility on identical instances and scorers.
For a held-out benchmark $b$, let $\mathcal{D}_b$ be its instance distribution, $c_b(\xi)$ the prompt constructed from instance $\xi$, and $s_b(\cdot,\xi)\in[0,1]$ its official scorer.
Benchmark utility on a percentage-point scale is
\begin{equation}
    U_b(f_\theta)=
    100\cdot
    \mathbb{E}_{\xi\sim\mathcal{D}_b}
    \left[s_b\!\left(f_\theta(c_b(\xi)),\xi\right)\right].
\end{equation}
The factor 100 converts the normalized mean score to percentage points.
For triggered utility $U_b(f_\theta\circ T_z)$, $z$ is inserted into the user-content field of $c_b(\xi)$; all other prompt construction and scoring remain unchanged.
The reported utility change is
\begin{equation}
    \label{eq:dutility}
    \dutility_b(z; f_\theta) =
    U_b(f_\theta\circ T_z)-U_b(f_\theta).
\end{equation}
Negative values indicate degradation.

\subsection{Clean Behavior as a Discovery Reference}
A usable source-side reference must be available for every unlabeled instruction, remain conditioned on that instruction, and avoid encoding any desired downstream failure.
The victim's clean response-onset preference provides such a reference: it is directly observable for each instruction and does not prescribe an alternative answer.
\method{} compares this preference under clean and span-modified versions of the same instruction.
Both passes use the same frozen $f_\theta$ and differ only in the inserted span, yielding an instruction-paired comparison.
This two-pass construction is analogous to shared-parameter comparisons in Siamese and contrastive learning \citep{bromley1993signature,wu2023autoaugmentation,muennighoff2025gritlm}, but compares response behavior rather than learned representations.

Let $\mathcal{U}$ denote the unlabeled source-instruction set.
For a clean instruction $u\in\mathcal{U}$, let $c_\theta(u)$ denote the victim-specific chat-formatted prompt and $p_\theta(\cdot\mid c_\theta(u))$ the next-token distribution at its end.
The clean preferred token at response onset is
\begin{equation}
    \hat{y}(u)=\arg\max_y p_\theta(y\mid c_\theta(u)).
\end{equation}
$\hat{y}(u)$ thus anchors the comparison to the victim's own clean behavior on that instruction.

After inserting $z$ at the chosen position, $T_z(u)$ is the modified source instruction and $c_\theta(T_z(u))$ applies the same chat template.
We quantify the candidate-induced change in the negative log-probability of this reference with Endogenous Preference Disruption (EPD):
\begin{align}
    \ell_0(u) &= -\log p_\theta(\hat{y}(u)\mid c_\theta(u)),\\
    \ell_z(u) &= -\log p_\theta(\hat{y}(u)\mid c_\theta(T_z(u))),\\
    r_{\mathrm{EPD}}(z;u) &= \ell_z(u)-\ell_0(u).
\end{align}
EPD is positive when insertion lowers the probability of the clean preferred token.
Subtracting $\ell_0$ expresses the score as a within-instruction change before scores from different instructions are pooled.
Higher EPD therefore means stronger source-local suppression of the clean preferred token; neither cross-input effectiveness nor downstream utility enters the score.
Whether this local disruption exposes a reusable vulnerability is tested only after selection, when the fixed trigger is reused across held-out inputs and tasks.

\subsection{Proposal Search and Fixed-Trigger Selection}
Testing this hypothesis requires searching a large discrete span space without introducing cross-input or downstream feedback.
For each victim, an auxiliary generator $g_\phi$ parameterizes a proposal distribution over bounded spans.
For every $u\in\mathcal{U}$, $g_\phi$ samples $z\sim g_\phi(\cdot\mid u)$ under a fixed meta-instruction specifying a short phrase and token budget $L$.
Each proposal is inserted into its originating source instruction and evaluated by the paired source-local score above.
We use Group Relative Policy Optimization (GRPO) to shift this proposal distribution toward spans with larger source-local scores; only $g_\phi$ is updated, while $f_\theta$ remains frozen \citep{shao2024deepseekmath}.

After source-side search for victim $v$, one draw $z_i\sim g_\phi(\cdot\mid u_i)$ per source instruction forms the indexed pool $\mathcal{C}_v=((u_i,z_i))_{i=1}^{N}$ for $\mathcal{U}=\{u_i\}_{i=1}^{N}$.
Repeated strings from different instructions are retained.
Candidates are ranked by EPD only on their originating instructions, so selection uses neither cross-input effectiveness nor downstream utility.
The selected trigger is
\begin{equation}
    z_v^\star=\arg\max_{(u,z)\in\mathcal{C}_v} r_{\mathrm{EPD}}(z;u).
\end{equation}
After selection, $g_\phi$ is no longer used and $z_v^\star$ is inserted unchanged into every held-out input for victim $v$; held-out utility change then tests cross-input and cross-task transfer.

\begin{table*}[htbp]
\centering
\setlength{\tabcolsep}{3pt}
\renewcommand{\arraystretch}{1.2}
\resizebox{\textwidth}{!}{%
\begin{tabular}{lrrrrrrrrrr}
\toprule
Method & MMLU & MMLU-Pro & GSM8K & MATH-500 & BBH & HSwag & ARC & TruthfulQA & IFEval & Avg. \\
\midrule
PromptBench & -2.3 (3.7\%) & -7.7 (17.6\%) & -7.6 (9.2\%) & -3.4 (8.9\%) & -14.2 (24.4\%) & -5.0 (6.9\%) & -8.3 (9.9\%) & -4.2 (8.8\%) & -5.9 (9.1\%) & -6.5 (10.6\%) \\
UAT & 2.0 (-4.5\%) & -0.2 (0.3\%) & -4.5 (7.9\%) & -1.1 (5.9\%) & -1.8 (3.6\%) & 5.1 (-12.0\%) & 0.3 (-0.5\%) & 5.3 (-62.8\%) & -1.3 (1.7\%) & 0.4 (-6.7\%) \\
GCG & -26.1 (44.5\%) & -15.9 (40.3\%) & -33.8 (48.1\%) & -12.2 (34.3\%) & -37.9 (66.2\%) & -25.5 (41.1\%) & -37.9 (45.9\%) & -12.8 (-4.0\%) & -16.7 (27.1\%) & -24.3 (38.2\%) \\
BEAST & 2.3 (-5.5\%) & -0.1 (-0.2\%) & -1.4 (1.6\%) & -1.4 (5.8\%) & -2.7 (4.6\%) & 5.9 (-13.3\%) & 0.4 (-0.5\%) & 0.6 (1.1\%) & -1.2 (2.1\%) & 0.3 (-0.5\%) \\
AmpleGCG & -0.1 (-0.4\%) & -0.5 (1.1\%) & -11.8 (21.3\%) & -1.2 (4.9\%) & -2.4 (4.5\%) & 3.8 (-10.0\%) & -1.1 (1.2\%) & 1.3 (-4.1\%) & -4.3 (7.1\%) & -1.8 (2.8\%) \\
AdvPrompter & -1.1 (1.8\%) & -6.1 (13.9\%) & -6.4 (7.8\%) & -2.6 (6.9\%) & -7.6 (13.0\%) & 1.6 (-2.2\%) & -1.7 (2.1\%) & -0.7 (1.6\%) & -5.7 (8.9\%) & -3.4 (5.5\%) \\
System Prompt Poisoning & -9.3 (12.4\%) & -3.4 (7.0\%) & -15.9 (21.4\%) & -8.8 (26.8\%) & -6.0 (8.9\%) & -37.5 (47.2\%) & -55.2 (63.3\%) & \textbf{-36.8 (62.0\%)} & 0.6 (-0.7\%) & -19.1 (27.6\%) \\
Hand-written & -1.4 (2.4\%) & -3.4 (7.7\%) & -34.3 (41.4\%) & -11.5 (30.3\%) & -31.5 (54.1\%) & -2.7 (3.8\%) & -1.1 (1.3\%) & -2.4 (5.1\%) & -20.3 (31.4\%) & -12.1 (19.7\%) \\
\midrule
\method{} (suffix) & \textbf{-36.2 (59.4\%)} & \textbf{-26.4 (59.9\%)} & \textbf{-58.4 (70.5\%)} & \textbf{-25.0 (65.8\%)} & \textbf{-43.6 (75.0\%)} & \textbf{-43.4 (60.0\%)} & \textbf{-62.4 (73.9\%)} & -33.2 (69.7\%) & \textbf{-33.1 (51.3\%)} & \textbf{-40.2 (65.4\%)} \\
\method{} (prefix) & -19.1 (31.4\%) & -2.9 (6.5\%) & -24.4 (29.5\%) & -19.6 (51.6\%) & -24.4 (42.0\%) & -35.1 (48.6\%) & -45.8 (54.2\%) & -28.1 (58.9\%) & -26.2 (40.6\%) & -25.1 (40.8\%) \\
\method{} (random) & -21.9 (35.9\%) & -9.6 (21.9\%) & -30.8 (37.1\%) & -18.1 (47.7\%) & -31.2 (53.7\%) & -33.6 (46.4\%) & -45.3 (53.7\%) & -25.1 (52.6\%) & -29.8 (46.1\%) & -27.3 (44.4\%) \\
\bottomrule
\end{tabular}
}
\caption{Utility change on held-out benchmarks. Entries are attacked-minus-clean percentage-point changes averaged over four victims; more negative values indicate greater degradation. Parentheses give the mean per-victim relative decrease, $100(U_{\mathrm{clean}}-U_{\mathrm{attack}})/U_{\mathrm{clean}}$. Avg. averages all 36 victim--benchmark cells, and boldface marks the greatest degradation in each column.}
\label{tab:main_results}
\end{table*}

\section{Experiments}
The experiments test whether source-local preference disruption identifies a reusable context vulnerability, then examine its dependence on insertion and reuse protocols, recovery across finite proposal supports, and downstream regression structure.

\subsection{Experimental Setup}
The source corpus is the instruction field of Databricks Dolly 15K; removing exact duplicates leaves 14,824 unlabeled instructions \citep{conover2023dolly}.
Training stops when smoothed EPD plateaus; the checkpoint with the highest logged training EPD produces one candidate per source instruction for selection.
Unless an ablation states otherwise, the auxiliary generator is Qwen3-1.7B \citep{qwen3technical} and the trigger budget is 32 generator tokens.
Candidate sampling uses temperature 1.0, top-$p$ 0.9, and repetition penalty 1.2.

Held-out evaluation covers four open-weight instruction-tuned victims: Qwen2.5-7B-Instruct, Llama-3.1-8B-Instruct, Mistral-7B-Instruct-v0.3, and Gemma-2-9B-IT \citep{qwen25technical,llama3herd,mistral7b,gemma2report}.
The nine benchmarks cover knowledge, mathematical reasoning, commonsense reasoning, truthfulness, and instruction following: MMLU \citep{hendrycks2020measuring}, MMLU-Pro \citep{wang2024mmlupro}, GSM8K \citep{cobbe2021training}, MATH-500 \citep{hendrycks2021measuringmath}, BBH \citep{suzgun2022challenging}, HellaSwag \citep{zellers2019hellaswag}, ARC \citep{clark2018arc}, TruthfulQA \citep{lin2022truthfulqa}, and IFEval \citep{zhou2023instruction}.
Each benchmark's native score serves as utility, reported as attacked-minus-clean change in percentage points.
On the common grid, each benchmark column averages over four victims and the overall mean covers all 36 victim--benchmark cells.

Each method is evaluated on the same victim--benchmark grid using its native objective, prompt channel, and final artifact.
UAT supplies a fixed prefix; GCG, BEAST, and AmpleGCG supply fixed suffixes; AdvPrompter generates an input-conditioned suffix.
The PromptBench pool combines character-, word-, sentence-, and semantic-level perturbations with punctuation and whitespace variants; it and a hand-written pool of verbosity, formatting, caution, and distraction instructions receive benchmark-specific oracle selection in every victim--benchmark cell.
System Prompt Poisoning provides a system-channel availability reference.
\method{} instead contributes one source-selected user suffix per victim, fixed before any benchmark score is observed and reused throughout the grid.

\subsection{Utility Transfer to Held-Out Tasks}
On the common evaluation grid in Table~\ref{tab:main_results}, the fixed \method{} suffix yields a mean $\dutility$ of -40.2 points.
The change is negative in 35 of the 36 individual victim--benchmark cells and in the four-victim mean for every benchmark.
Discovery uses only a clean response-onset preference on unlabeled source instructions, whereas the held-out benchmarks score complete outputs with task-specific criteria.
Because neither discovery nor selection observes benchmark feedback, this 35-of-36 pattern provides evidence that source-selected spans produce utility degradation that transfers across inputs and tasks rather than reflecting direct optimization of the measured outcomes.

Even with benchmark-specific oracle selection, PromptBench and the hand-written controls average -6.5 and -12.1 points.
GCG is the strongest fixed-user-suffix baseline at -24.3, while System Prompt Poisoning averages -19.1 and exceeds \method{} only on TruthfulQA.
The evaluated generic perturbations, jailbreak-optimized suffixes, and privileged-channel poisoning impair benign utility without matching \method{}'s mean degradation under the reported protocols.

We next ask whether source-local discovery is tied to suffix insertion or also recovers transferable spans at other positions.
Repeating discovery, selection, and evaluation with prefix or random-position insertion yields mean $\dutility$ values of -25.1 and -27.3 points, respectively, compared with -40.2 for suffix insertion.
Because discovery and selection are repeated at each location, these rows compare complete position-specific protocols rather than moving one trigger between positions.
Transfer is therefore not confined to suffix insertion, although its observed magnitude varies across position-specific protocols and is largest for the suffix setting.

\subsection{Protocol and Selection Ablations}
Table~\ref{tab:input_length} separates two questions: whether stronger transfer depends on source-side selection and whether the default result is explained by token budget.
For the first, the null-context control reuses one trigger generated from a fixed placeholder, whereas per-input generation conditions a new trigger on each evaluation prompt.
Both use prefix insertion and are compared with the source-selected fixed-prefix protocol, which yields -25.1 points in Table~\ref{tab:main_results}.
The null-context and per-input controls yield -16.26 and -14.33 points, respectively.
This matched comparison associates the stronger result with selection over unlabeled source instructions rather than fixed reuse alone or input-specific conditioning; the selected span then transfers without adaptation.

For length, each token budget defines a separate fixed-suffix search.
The 16-token four-victim run yields -20.89 points and the 64-token Qwen run -26.12, versus their 32-token references of -40.20 and -66.5 points, respectively.
Across these runs, degradation is non-monotonic in budget, arguing against a length-only account.

\begin{table}[t]
\centering
\scriptsize
\setlength{\tabcolsep}{5pt}
\renewcommand{\arraystretch}{1.2}
\begin{tabular}{lc}
\toprule
Setting & Mean $\dutility$ \\
\midrule
Null-context fixed prefix, 32 tokens & -16.26 \\
Per-input prefix generation, 32 tokens & -14.33 \\
Source-selected fixed suffix, 16 tokens & -20.89 \\
Qwen-only fixed suffix, 64 tokens & -26.12 \\
\textbf{Source-selected fixed suffix, 32 tokens} & \textbf{-40.20} \\
\bottomrule
\end{tabular}
\caption{Protocol and length controls. Values are mean $\dutility$ in percentage points; more negative values indicate greater degradation. The first two rows are prefix controls, with the matched source-selected prefix reported in Table~\ref{tab:main_results}; the remaining rows use fixed suffixes. Rows average four victims and nine benchmarks except the Qwen-only 64-token diagnostic.}
\label{tab:input_length}
\end{table}

We next test whether transfer depends on the single highest-EPD candidate or extends across the retained high-EPD pool.
We first characterize realized proposal support by tracking generated strings on 1,000 fixed Dolly prompts for each default-generator run, merging strings after lowercasing and whitespace normalization.
Normalized Shannon effective support is the exponentiated normalized-string entropy divided by 1,000.
From the first to the last logged checkpoint, this support falls from 0.920--1.000 to 0.007--0.063, while Top-10 normalized-string mass rises from 0.010--0.040 to 0.559--0.926.
These trajectories show that realized proposal support can contract sharply during training.
Related GRPO work associates such concentration with reduced exploration, while adversarial-suffix work finds successful candidates beyond the single lowest-loss suffix \citep{yu2025dapo,liao2024amplegcg}.
We therefore test selection breadth directly in the retained checkpoint's realized high-EPD pool: Table~\ref{tab:selection_calibration} asks whether transfer extends beyond Top-1 among unique candidates.

\begin{table}[t]
\centering
\scriptsize
\setlength{\tabcolsep}{0pt}
\renewcommand{\arraystretch}{1.02}
\begin{tabular*}{\linewidth}{@{\extracolsep{\fill}}lrrrrrrrr@{}}
\toprule
& \multicolumn{2}{c}{Top-1} & \multicolumn{3}{c}{\shortstack{Source EPD\\retained}} & \multicolumn{3}{c}{\shortstack{Utility loss\\retained}} \\
\cmidrule(lr){2-3}\cmidrule(lr){4-6}\cmidrule(lr){7-9}
Victim & EPD & $\Delta U$ & @5 & @10 & @20 & @5 & @10 & @20 \\
\midrule
Qwen2.5-7B & 62.1 & -66.5 & 97.1 & 95.1 & 92.9 & 99.7 & 99.4 & 98.5 \\
Llama-3.1-8B & 33.4 & -20.5 & 96.7 & 90.6 & 82.8 & 87.4 & 86.4 & 84.5 \\
Mistral-7B & 35.2 & -27.9 & 95.8 & 94.2 & 91.4 & 101.5 & 101.4 & 100.0 \\
Gemma-2-9B & 45.8 & -45.9 & 100.0 & 100.0 & 99.0 & 114.9 & 111.0 & 111.0 \\
\bottomrule
\end{tabular*}
\caption{Breadth of source-ranked transfer. $\Delta U$ is mean attacked-minus-clean utility over nine held-out benchmarks for each victim; negative values indicate loss. Retention@$K$ averages the first $K$ unique candidates in the source-EPD ranking and divides their mean source EPD or utility loss $-\Delta U$ by the corresponding Top-1 value (\%).}
\label{tab:selection_calibration}
\vspace{-0.3cm}
\end{table}

\begin{figure*}[htbp]
\centering
\includegraphics[width=0.98\textwidth]{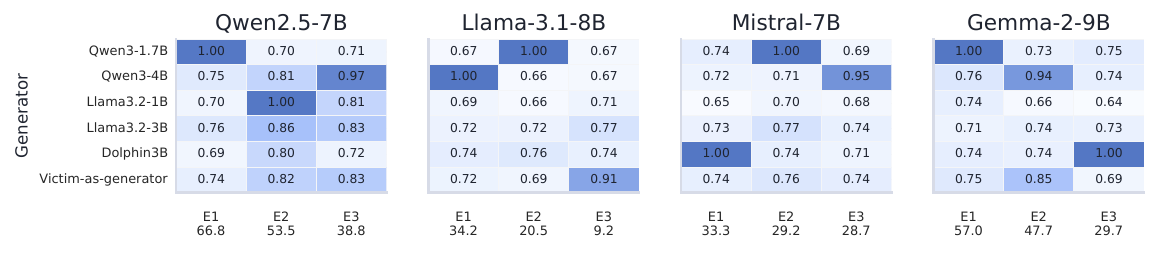}
\vspace{-0.4cm}
\caption{Realized-pool proximity to post-hoc high-degradation anchors. For each victim, E1--E3 are selected by held-out degradation from the 30 highest-EPD unique candidates per proposer, subject to pairwise Sentence-T5 cosine below 0.85. Cells report each pool's frequency-weighted 95th-percentile cosine; values below the anchors report utility loss $-\Delta U$ in points.}
\label{fig:effective_family_coverage}
\vspace{-0.2cm}
\end{figure*}

All 80 unique candidates in the four source-ranked Top-20 lists have negative mean held-out $\Delta U$.
At $K=20$, their mean EPD retains 82.8--99.0\% of Top-1 and their mean utility loss retains 84.5--111.0\%.
For Mistral and Gemma, the top-five mean utility loss exceeds the Top-1 loss, despite downstream utility never entering the source ranking.
Source-side ranking therefore identifies a transferable candidate set rather than an isolated optimum, while downstream evaluation supplies a distinct severity ordering only after discovery.

\subsection{Transfer Across Proposal Models}
Within-pool breadth does not establish whether source-only discovery continues to recover transferable spans when the proposal distribution changes.
A proposal-based search can score and select only candidates that its proposer realizes; we therefore repeat the complete discovery protocol with Qwen3-1.7B and five alternative proposal backbones.
For each victim--generator pair, one fixed trigger is selected while the remaining discovery choices and the nine-benchmark evaluation grid are held fixed.

\begin{table}[htbp]
\centering
\scriptsize
\setlength{\tabcolsep}{2.2pt}
\renewcommand{\arraystretch}{1}
\begin{tabular}{lccccc}
\toprule
Generator & \shortstack{Qwen2.5-7B} & \shortstack{Llama-3.1-8B} & \shortstack{Mistral-7B} & \shortstack{Gemma-2-9B} & Mean \\
\midrule
Qwen3-1.7B & -66.5 & -20.5 & -27.9 & -45.9 & -40.2 \\
Qwen3-4B & -10.4 & -34.2 & -28.7 & -47.7 & -30.3 \\
Llama-3.2-1B & -30.2 & -5.0 & -10.1 & -1.1 & -11.6 \\
Llama-3.2-3B & -8.5 & -3.4 & -23.3 & -12.4 & -11.9 \\
Dolphin3.0-3B & -10.0 & -2.7 & -33.3 & -29.7 & -18.9 \\
Victim-as-generator & -12.9 & -9.2 & -25.6 & -47.0 & -23.7 \\
\bottomrule
\end{tabular}
\caption{Auxiliary-generator comparison for fixed-trigger transfer. Each cell reports mean $\dutility$ across nine benchmarks for the fixed trigger selected for one victim--generator pair. Mean is the unweighted average of the four victim columns; more negative values indicate greater degradation.}
\label{tab:generator_family}
\end{table}

All 24 victim--generator pairs yield negative mean utility change, showing that the direction of transfer is not peculiar to the default proposer.
At the same time, the 65.4-point span across pairs shows that recurrence of transfer does not imply uniform recovery strength.
The strongest proposer changes with the victim.
Neither increasing generator scale within the Qwen3 and Llama-3.2 pairs nor using the victim itself consistently strengthens degradation, so these simple proposer attributes provide no stable ordering across the realized pools.

The sweep establishes variation in recovered strength but does not show how the realized candidate pools differ.
As a retrospective diagnostic of realized search support, Figure~\ref{fig:effective_family_coverage} measures each pool's proximity in Sentence-T5 space to three high-degradation anchors selected post hoc from completed Top-30 evaluations \citep{ni2022sentencet5}.
For every victim, the three anchors originate from different proposers, while the other proposer pools' frequency-weighted 95th-percentile similarities to them span 0.64--0.97.
The anchors are therefore not confined to one proposer, and the realized pools reach their embedding neighborhoods to different degrees.
The sweep and geometric diagnostic support a finite-search interpretation: the strength recovered for a victim is a joint property of its context vulnerability and the candidate neighborhoods exposed by the proposer.

\subsection{Structure and Semantic Forms of Regressions}
\begin{figure}[htbp]
\centering
\includegraphics[width=\columnwidth,trim=0 0 291.45 0,clip]{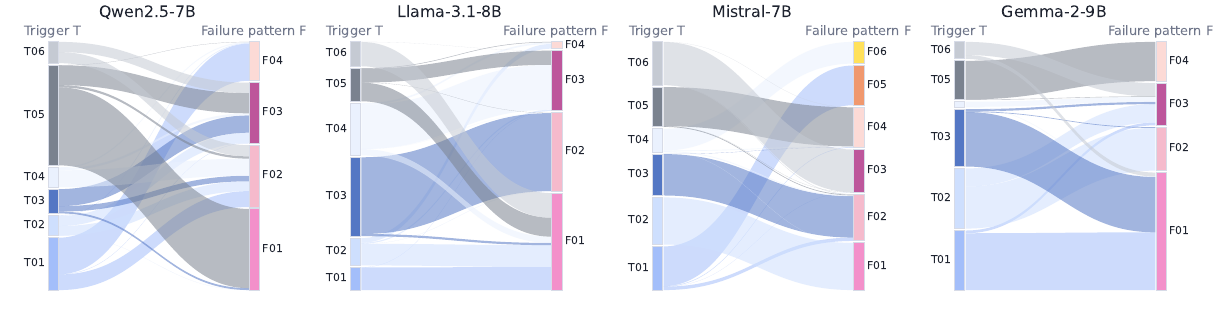}
\includegraphics[width=\columnwidth,trim=291.45 0 0 0,clip]{figures/generator_family_failure_flow.pdf}
\vspace{-0.4cm}
\caption{Victim-local trigger clusters ($Txx$) and attacked-output clusters ($Fxx$) for the evaluated fixed suffixes; ribbon width counts paired regressions. Labels are nonsemantic and local to each victim; $Fxx$ is ordered by cluster size within each panel, so $F01$ is not shared across victims.}
\label{fig:generator_failure_flow}
\vspace{-0.2cm}
\end{figure}

Beyond recurrence and recovered strength, we ask whether regressions from each fixed victim--trigger pair are diffuse or concentrated in a recurring output region.
Because EPD specifies no task-level failure form, aggregate utility loss alone cannot answer this question.
For each victim, Sentence-T5 embeddings of candidates from all six proposers are clustered with generation-frequency weights, defining a common victim-local candidate space.
Its $Txx$ clusters partition the realized pools without using held-out degradation.
Attacked outputs are clustered separately as $Fxx$ for paired regressions, defined as inputs with a positive clean score and a lower triggered score under the same task scorer.
Frequency-weighted $k$-means is fit over candidate and output cluster counts $\{6,8,10,12,16,20\}$ and $\{4,5,6,7\}$, respectively, with cosine silhouette selecting $k$.
Figure~\ref{fig:generator_failure_flow} links each evaluated fixed suffix's $Txx$ cluster to its $Fxx$ clusters across 150--2,237 regressions per victim--trigger pair.
At least 70\% of regressions occupy one output cluster for 19 of the 24 pairs, and 13 exceed 90\%.
By contrast, after pooling all six triggers for each victim, the largest output cluster accounts for only 19.8--48.5\%.
This contrast shows that recurrence is concentrated within victim--trigger pairs rather than inherited from the pooled victim-level output distribution, even though discovery specifies no output region.

To assign shared semantics to these victim-local patterns, three annotators independently review 1,000 benchmark-stratified regressions from the main protocol, 250 per victim, using a six-category codebook.
Records contain the prompt, reference, and paired outputs but omit the victim identity and trigger text; labels require unanimous agreement, with disagreements independently re-annotated.

\begin{figure}[htbp]
\centering
\includegraphics[width=0.98\linewidth]{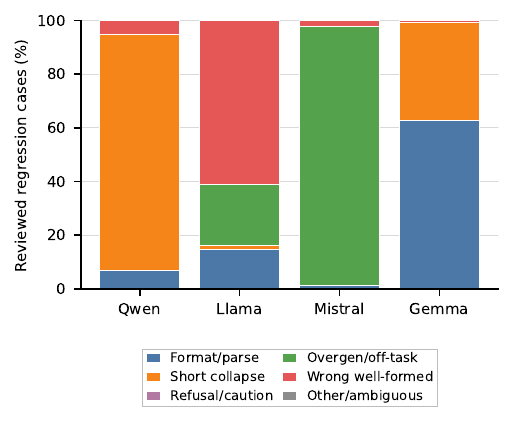}
\vspace{-0.4cm}
\caption{Failure categories in 1,000 benchmark-stratified paired regressions (250 per victim). The same six-category codebook is used across victims; percentages are conditional on the reviewed regressions, not on all benchmark inputs.}
\label{fig:human_failure_attribution}
\vspace{-0.3cm}
\end{figure}

The dominant category differs sharply across the four reviewed pairs (Figure~\ref{fig:human_failure_attribution}): short collapse accounts for 88.0\% of Qwen regressions, whereas wrong but well-formed answers account for 61.2\% of Llama regressions.
Off-task over-generation accounts for 96.8\% of Mistral regressions, whereas Gemma regressions divide between format/parse failure (62.8\%) and short collapse (36.4\%).
The clustering and semantic review therefore reveal structured heterogeneity: regressions recur within evaluated victim--trigger pairs, but their dominant form is neither shared across pairs nor specified by the discovery objective.
Together with the breadth and proposer results, these analyses separate three empirical properties: the presence of reusable context vulnerabilities, their visibility under finite proposal search, and the victim--trigger-specific forms through which they manifest downstream.

\section{Limitations}
Our experiments establish breadth within realized proposal pools, not coverage of the full bounded-span space.
Generated strings become increasingly concentrated during training, and alternative proposers recover substantially different held-out degradation under the same victim and EPD criterion.
These findings make proposal coverage part of the discovery problem: a negative search outcome may reflect insufficient exploration rather than the absence of a transferable span.
Future work should develop diversity-preserving search and coverage diagnostics that quantify how the set of recovered transferable spans expands as proposal support broadens.

The clustering and annotation analyses reveal recurrent, pair-specific failure manifestations, but not the causal computations linking response-onset disruption to downstream generation.
Cross-trigger causal interventions on internal activations could test whether these regressions share a common pathway within the victim model or arise through distinct mechanisms.
The present implementation also requires next-token probabilities from an open-weight victim model; extending endogenous discovery to restricted-access models remains open.

\section{Conclusion}
When downstream failure is unspecified and downstream task feedback is withheld during discovery, a frozen model's clean behavior can reveal fixed spans that transfer across unseen inputs and tasks.
Across four victims and nine benchmarks, \method{} yields a 40.2-point mean utility loss, the largest among evaluated methods under the reported protocols.
Transfer recurs across ranked candidates and proposal models, but strength varies across realized proposal pools: broad recurrence does not imply exhaustive search coverage.
In 19 of 24 victim--trigger pairs, regressions are far more concentrated than the corresponding victim-level marginals, indicating that fixed spans repeatedly induce restricted output regimes rather than diffuse degradation.
Together, these results distinguish vulnerability existence, finite-search visibility, and victim--trigger-specific manifestation; aggregate utility captures severity but not this structure.
Clean model behavior can thus serve as an endogenous discovery signal, making downstream failure a manifestation to characterize rather than a target to prescribe.

\section*{Ethics Statement}
This work studies a dual-use attack on the benign utility of instruction-tuned LLMs.
Evaluations use open-weight models and public benign benchmarks; no experiment targets private data, deployed services, or users.
The findings can support detection and mitigation.
We will release training and evaluation code, configurations, trigger hashes, and analysis tools, while sharing raw triggers and generator weights only with vetted researchers under responsible disclosure.


\bibliography{main}

\end{document}